\documentclass[sigconf,10pt]{acmart}
\setcopyright{none}
\renewcommand\footnotetextcopyrightpermission[1]{}
\usepackage{tikz}
\usepackage{xcolor}
\usepackage{graphicx}
\usepackage{caption}
\usepackage{subcaption}
\usepackage{booktabs}
\usepackage{microtype}          
\usepackage[htt]{hyphenat}      
\usepackage{xspace}             
\usepackage{pifont}             

\newcommand{\ourmethod}{\textsc{Plume}\xspace}

\newenvironment{sitemize}
  {\begin{itemize}
      \setlength{\itemsep}{0.3em}
      \setlength{\topsep}{0pt}
  }
  {\end{itemize}}

\begin{document}

\title{\ourmethod: Building a Network-Native Foundation Model for Wireless Traces via Protocol-Aware Tokenization}

\author{Swadhin Pradhan}
\email{swapradh@cisco.com}
\affiliation{%
  \institution{Cisco Systems}
  \country{USA}
}

\author{Shazal Irshad}
\email{sirshad@cisco.com}
\affiliation{%
  \institution{Cisco Systems}
  \country{USA}
}

\author{Jerome Henry}
\email{jerhenry@cisco.com}
\affiliation{%
  \institution{Cisco Systems}
  \country{USA}
}

\begin{abstract}
Foundation models succeed when they learn in the native structure of a modality, whether morphology-respecting tokens in language or pixels in vision.
Wireless packet traces deserve the same treatment: meaning emerges from layered headers, typed fields, timing gaps, and cross-packet state machines, not flat~strings.
We present \textbf{\ourmethod{}} (\textbf{P}rotocol \textbf{L}anguage \textbf{U}nderstanding \textbf{M}odel for \textbf{E}xchanges), a compact 140M-parameter foundation model for 802.11 traces that learns from structured PDML dissections.
A \emph{protocol-aware tokenizer} splits along the dissector field tree, emits gap tokens for timing, and normalizes identifiers, yielding $6.2\times$ shorter sequences than BPE with higher per token information density.
Trained on a curated corpus, \ourmethod{} achieves 74--97\% next-packet token accuracy across five real-world failure categories and AUROC$\geq$0.99 for zero-shot anomaly detection.
On the same prediction task, frontier LLMs (Claude Opus~4.6~\cite{anthropic_claude46_2025}, GPT-5.4~\cite{openai_gpt54_2025}) score comparably despite receiving identical protocol context, yet \ourmethod{} does so with $>$600$\times$ fewer parameters, fitting on a single GPU at effectively zero marginal cost vs.\ cloud API pricing, enabling on-prem, privacy-preserving root cause analysis.
\end{abstract}

%
%

\maketitle
\pagestyle{plain}
\fancyhead{}
\renewcommand{\headrulewidth}{0pt}

\section{Introduction}
\label{sec:intro}

Foundation models succeed when they learn in the native structure of a modality.
In language, GPT-4 and PaLM learn from tokens respecting morphology~\cite{openai_gpt4_2023,chowdhery2022palm}; in vision, modern backbones learn directly from pixels.
Wireless networking deserves the same treatment.
802.11 packet exchanges are not free-form text: meaning emerges from layered headers, Information Elements (IEs) carrying negotiated options, timing gaps, and cross-packet state-machine~transitions.
Generalist Large Language Models (LLMs) that see packets as flattened strings rarely internalize this structure.

\textbf{\ourmethod} (\textbf{P}rotocol \textbf{L}anguage \textbf{U}nderstanding \textbf{M}odel for \textbf{E}xchanges) is a compact, \emph{network-language-native} foundation model for 802.11 wireless traces.
It learns from \emph{structured dissections}, specifically Wireshark / tshark Packet Description Markup Language (PDML) exports~\cite{wireshark_pdml_wiki,tshark_manual,wireshark_xml_readme}.
Rather than claiming novelty in pretraining over network data (cf.\ Lens~\cite{wang2024lens}, netFound~\cite{guthula2023netfound}), we push toward a design where \emph{representation, tokenization, and data quality} are first-class levers, and outputs can be \emph{natural-languagified} to interface with LLM planners and chat agents for Root Cause Analysis (RCA)~\cite{lewis2020rag}.

\textit{Why not simply fine-tune a general LLM?}
Fine-tuning on text tokens preserves an interface mismatch.
The model sees surface strings rather than typed protocol fields or state-machine transitions, encouraging shortcuts that mimic reasoning but encode bias rooted in the tokenized surface, not the protocol~itself.
Distillation compounds the problem by compressing the same shortcuts while shedding capacity to question them.
We validate this empirically: frontier LLMs (Claude Opus~4.6~\cite{anthropic_claude46_2025}, GPT-5.4~\cite{openai_gpt54_2025}) given identical protocol context achieve 79--94\% token accuracy on next-packet prediction, while \ourmethod{} reaches 75--96\% with a $>$600$\times$ smaller model (\S\ref{sec:eval:llm}).
A packet-native model amortizes learning once, capturing reusable structure for association or authentication dialogues, control--data interleavings, error signatures, and airtime~patterns.
Practically, a compact specialized model is simpler to serve on-prem, respects privacy by avoiding external APIs, and fits as a callable tool in multi-agent workflows, aligning with compute-optimal lessons that advocate scaling tokens and model size in tandem~\cite{hoffmann2022chinchilla}.

\subsection{Representation and Tokenization}

\textbf{Structured dissections as anchors.}
We treat PDML exports as primary training substrates because they preserve the protocol tree (typed fields, byte offsets, parent--child relations) while remaining integrable with scrubbing, annotation, and labeling pipelines.
PDML provides stable field names (e.g., \texttt{wlan.fc.\allowbreak{}type\_subtype}, \texttt{rsn.capabilities}), explicit typing, and consistent hierarchy~\cite{wireshark_pdml_wiki,tshark_manual,wireshark_xml_readme,wireshark_export_json}.
Raw PCAPs remain first-class for re-dissection, but structured views anchor learning at the right boundaries.
Prior foundation-style traffic models~\cite{wang2024lens,guthula2023netfound} demonstrate the promise of pretraining over network data; \ourmethod{} complements them by centering tokenizer design on field and timing semantics and by supporting a natural-language surface for~interoperability.

\textbf{Network-language-native tokenization.}
Tokenization is among the most impactful design choices.
Off-the-shelf Byte Pair Encoding (BPE)~\cite{sennrich2016bpe} or fixed byte chunks smear field boundaries and timing, yielding long, low-signal sequences; our experiments confirm BPE on PDML yields $6.2\times$ more tokens per packet than \ourmethod's field-value tokenizer (Table~\ref{tab:tokenization_ablation}).
We adopt \emph{protocol- and timing-aware} tokenization that
(i)~splits along the dissector field tree,
(ii)~emits \emph{gap tokens} for inter-arrival times and aggregation windows,
(iii)~normalizes identifiers that should be compared structurally rather than memorized (MACs, SSIDs, etc.),
(iv)~adaptively sub-tokenizes variable-length options, and
(v)~optionally attaches \emph{natural-languagified glosses} so outputs are legible to general LLMs without exposing raw payloads.
This aligns with recent byte-level modeling showing that dynamic, content-aware patching can rival fixed vocabularies at scale~\cite{pagnoni2024blt}. We pair this tokenizer with a GPT-style~\cite{radford2019gpt2} auto-regressive objective that models the packet conversation across time.

\subsection{Data Quality}

Scaling tokens alone is not enough; \emph{what} you pretrain on matters as much as \emph{how much}.
Na\"ive ``capture everything'' produces severe skew: today's pipelines are largely \emph{reactive}, so by the time an alert fires, pre-failure context (TCP SYN/ACK, DHCP DISCOVER/OFFER, 802.11 auth/assoc) is gone.
Datasets over-represent failures, under-represent healthy baselines, and rarely align positives and negatives under the same SSID, or RF~channel, yielding fragile classifiers.

We address this via \emph{proactive intelligent capture} that limits data explosion while preserving what teaches structure: edge agents with first-sign-of-life buffers maintain rolling pre-trigger windows; an adaptive positive/negative sampling engine constructs matched cohorts under identical RF/policy contexts; and Context-Enriched Capture Bundles~(CECBs) pair PCAPs with synchronized metadata (Received Signal Strength Indicator (RSSI) / Channel State Information (CSI) summaries, AP firmware, congestion counters, policy~state).

Our curation pipeline uses Hierarchical Density-Based Spatial Clustering (HDBSCAN)~\cite{mcinnes2017hdbscan} and Maximal Marginal Relevance (MMR)~\cite{carbonell1998mmr} sampling to reduce beacon dominance from $>$50\% to 4.7\% in the training set while preserving high per-token entropy (Table~\ref{tab:dataset_stats}), aligning with evidence that quality trumps raw token count~\cite{hoffmann2022chinchilla,lee2022dedup,penedo2023refinedweb,xie2023doremi}.

\subsection{From Model to System: Toolability}

\ourmethod{} is designed to be \emph{callable}.
A planner (LLM or rule engine) passes \ourmethod{} a PDML slice for a suspect interval; \ourmethod{}~returns
(i)~a structured summary at flow and packet levels,
(ii)~inconsistency and wrong-field flags, and
(iii)~localized hypotheses (e.g., ``PMF mismatch with legacy STA,'' ``PS-mode buffering $\rightarrow$ latency spikes'').
At 140M parameters it runs on-prem near capture points, exchanges \emph{explanations} rather than raw packets, and respects strict data-residency constraints, processing $\sim$200~packets/sec on a single NVIDIA A10G at effectively zero marginal cost (Table~\ref{tab:efficiency}).

\subsection{Contributions}

\begin{enumerate}
\item \textbf{\ourmethod}, a compact 140M-parameter foundation model for wireless traces that learns from structured PDML dissections with byte-level fallback, aligning inductive bias with protocol semantics.

\item A \textit{protocol- and timing-aware tokenizer} with field-boundary splits, gap tokens, identifier normalization, and adaptive IE segmentation, producing $6.2\times$ shorter sequences than BPE (Table~\ref{tab:tokenization_ablation}).

\item An HDBSCAN~\cite{mcinnes2017hdbscan}+MMR~\cite{carbonell1998mmr} curation pipeline reducing beacon bias from $>$50\% to 4.7\% while preserving rare events~\cite{hoffmann2022chinchilla,lee2022dedup,penedo2023refinedweb,xie2023doremi}.

\item Evaluation on five real-world 802.11 failure categories (50 PCAPs each) showing 74.1--97.3\% token accuracy, zero-shot AUROC$\geq$0.99 for anomaly detection, and 73.2\% five-class root cause accuracy from unsupervised features, and a head-to-head comparison with frontier LLMs showing that \ourmethod{} matches or exceeds both on next-packet prediction with $>$600$\times$ fewer parameters (Table~\ref{tab:llm_comparison}).

\item \textit{System integration}: \ourmethod{} as a callable tool in multi-agent RCA, enabling on-prem, privacy-preserving deployments that exchange structured explanations instead of raw packets.
\end{enumerate}
\section{Motivation and Background}
\label{sec:motivation}

Enterprise wireless networks generate vast volumes of packet traces rich in diagnostic information, from authentication handshakes and association sequences to EAPOL exchanges and data~flows.
When failures occur (e.g., bad passwords, EAPOL timeouts, invalid Pairwise Master Key Identifiers (PMKIDs)), the root cause is typically buried in subtle cross-packet patterns such as a missing acknowledgment, an unexpected field value, or an anomalous timing~gap.
Diagnosing these failures demands deep protocol expertise and manual Wireshark inspection, a process that does not scale to modern deployments with hundreds of sites and thousands of~clients.

\textbf{The promise and limits of general LLMs.}
LLMs have shown remarkable capability, yet applying them directly to packet analysis faces five obstacles.
(1)~Packets are structured, typed, hierarchical protocol data, not natural language;
(2)~standard tokenizers (BPE~\cite{sennrich2016bpe}, byte-level) destroy field boundaries, yielding sequences $6$--$16\times$ longer than necessary (Table~\ref{tab:tokenization_ablation});
(3)~sending raw packet data to cloud APIs raises privacy and compliance concerns (GDPR, HIPAA, data-residency);
(4)~API costs scale linearly with volume; we estimate \$4.92 per 1K~packets for GPT-5.2~\cite{openai_gpt52_2025} (Table~\ref{tab:efficiency}); and
(5)~even frontier models (Claude Opus~4.6, GPT-5.4) achieve only 86--89\% token accuracy on next-packet prediction, comparable to a 140M-parameter protocol-native model (\S\ref{sec:eval:llm}).

\textbf{The data quality gap.}
Existing packet datasets suffer from severe imbalance.
In 802.11 networks, APs transmit beacons every $\sim$102\,ms across multiple SSIDs and bands, so a 10-minute capture can contain $>$100K quasi-identical beacons dominating any na\"ive training~set.
Failure-mode captures are reactive, triggered after the fact, so pre-failure context (initial handshakes, setup packets) is often~missing.
Models trained on such skewed data learn beacon statistics rather than protocol dynamics.

\textbf{Our approach.}
\ourmethod{} addresses these gaps through three interlocking design choices:
(1)~a protocol-aware tokenizer respecting field boundaries and protocol hierarchy, yielding $6.2\times$ shorter sequences than BPE;
(2)~a curated training corpus where HDBSCAN~\cite{mcinnes2017hdbscan} clustering and MMR~\cite{carbonell1998mmr} sampling eliminate redundancy while preserving rare events, reducing beacon dominance from $>$50\% to 4.7\%; and
(3)~a family of compact auto-regressive architectures (140M--450M parameters) that run on-prem, enabling privacy-preserving deployment as a callable tool in multi-agent RCA~workflows.

\section{Tokenization for Network Captures}
\label{sec:tokenization}

The token is the fundamental semantic unit upon which the entire system is built. It is not merely a vocabulary-reduction device, but the determinant of what the model can learn, how efficiently it learns, and how far it can see within a fixed context~window. Traditional tokenizers fail for network captures because they ignore protocol hierarchy; \ourmethod's protocol-aware tokenizer addresses each failure~mode.

\subsection{Why Traditional Tokenizers Fail}
\label{sec:tokenization:why}

Traditional tokenizers such as BPE~\cite{sennrich2016bpe} seek short, reusable sub-word units.
This makes sense for human languages, where
\textcolor{red}{speak}%
\textcolor{blue}{er}
and
\textcolor{red}{speak}%
\textcolor{green}{ing}
share the root \textcolor{red}{speak}, and the suffixes \textcolor{blue}{er} and \textcolor{green}{ing} transfer across many roots.
However, this principle breaks down for network~captures.

Packets are ordered bit series, each position encoding a specific role (source address, upper-layer protocol, etc.).
Dissection tools such as Wireshark translate these into field names, e.g., \texttt{wlan.da} for the 802.11 destination address and \texttt{wlan.fc.type\_subtype} for the frame type and~subtype.
Confronted with these, BPE discovers sub-words like \emph{lan} and \emph{wlan}, producing tokens such as \emph{w}, \emph{lan}, \emph{sub}, and \emph{type}, yielding sequences of the~form:
\textcolor{red}{w}%
\textcolor{orange}{lan.}%
\textcolor{purple}{fc.}%
\textcolor{blue}{type}%
\textcolor{violet}{\_}%
\textcolor{lime}{sub}%
\textcolor{green}{type}%
\textcolor{black}{=}%
\textcolor{magenta}{0}%
\textcolor{cyan}{x}%
\textcolor{brown}{000}%
\textcolor{teal}{8}.

The resulting fragmentation is problematic in two ways.
First, orphan tokens like \emph{\_} or \emph{x} carry no semantic content yet consume context-window budget.
Second, the sub-word decomposition encodes linguistic relationships (e.g., \emph{WLAN} as a type of \emph{LAN}, \emph{subtype} as a subclass of \emph{type}) irrelevant to protocol~analysis: the fact that \emph{subtype} is linguistically subordinate to \emph{type} has no bearing on the values these fields carry; they could be called A and B with identical diagnostic~utility.
Together, these pathologies inflate sequences by $6.2\times$ relative to field-level tokenization (Table~\ref{tab:tokenization_ablation}), starving the model of the cross-packet context needed to learn protocol~dynamics.

\subsection{A Protocol-Aware Tokenizer}
\label{sec:tokenization:design}

We design a tokenizer aligned with the structure of captures rather than that of English.
Our first design choice is to assign one token per field name, so that \texttt{wlan.fc.type\_subtype} is a single, atomic token.
This is not entirely new; netFound~\cite{guthula2023netfound} and DBF-PSR~\cite{Ding2025DBF-PSR} adopt similar field-level tokenization, but our treatment of field \emph{values} is where the design~diverges.

Fields carry values of three distinct types (strings, symbols, and numerical quantities), and we handle each differently.

\textbf{Strings.}
Most strings are upper-layer (Layer~7) entries meaningful in the human dimension, such as a URL, an application name, or an HTTP user~agent.
We tokenize these with a secondary BPE tokenizer adapted to human language, preserving their natural sub-word~structure.

\textbf{Symbols.}
\ourmethod{} could learn to recognize codes irrespective of representation, e.g.,
\texttt{tcp.flags\allowbreak{}=\allowbreak{}0x10}
versus
\texttt{tcp.flags\allowbreak{}=\allowbreak{}ACK}.
However, network exchanges are dialogues, and symbols mark their~rhythm.
A TCP ACK validates that the previous segment was received, just as an 802.11 ACK validates reception of the prior frame.
This semantic equivalence is opaque with hex codes
(\texttt{0x10}, \texttt{0x1D})
but immediately visible with symbolic names
(\texttt{ACK}).
We therefore expand all symbols to their word representation.

\textbf{Numerical values.}
Network captures carry a wide variety of numerical values: time-series quantities (\texttt{frame.time\_delta}), identifiers (\texttt{ip.src\,=\,192.168.0.2}), and measurements (\texttt{wlan\_radio.\allowbreak{}signal\_dbm\,=\,-62}).
Measurements express both a quantity (``$-62$~dBm'') and a quality (``signal level is good'').
We retain raw numerical values during pretraining and teach the model to associate quantity ranges with qualitative meaning during post-training.\looseness=-1

Identifiers require special care because they convey multiple levels of~meaning.
The address \texttt{192.168.0.2} identifies a unique device, but an administrator also knows it is the second host in the \texttt{192.168.0.0/24} subnet, that any \texttt{192.168.0.x} address shares the same Layer~2 domain, and that the domain contains at most 254~hosts.
We represent IP addresses in two complementary forms during pretraining: as a string capturing device identity, and as a group of numbers capturing the hierarchical address structure.
We apply the same dual representation to MAC addresses: the vendor OUI is separated from the device-specific suffix, letting the model learn vendor-specific patterns.

\subsection{Field Filters and Layer Identifiers}
\label{sec:tokenization:filters}

A frame in a network capture is a long series of fields and values, many redundant or irrelevant.
Several fields express the same quantity in different contexts (e.g., \texttt{radiotap.\allowbreak{}dbm\_antsignal}, and \texttt{wlan\_radio.\allowbreak{}signal\_dbm} all report received signal~strength).
We remove such duplicates and suppress fields carrying only vacuous negative flags; for example, a 5\,GHz capture that reports ``not 900\,MHz,'' ``not 800\,MHz,'' ``non-CCK,'' and ``non-GSM.''
Out of 100--120 fields per frame, this suppresses $\sim$40, retaining only fields carrying positive information or negative information where the positive case is~plausible.

Network captures also encode protocol layering, visible in Wireshark's tree view or in PDML's hierarchical~structure.
When converted to a flat sequence, this layering is~lost.
However, layering is fundamental: the model must distinguish an 802.11 Layer~2 ACK from a TCP Layer~4 ACK, because these flags express dialogues between different entities susceptible to different failure~modes.
We therefore insert explicit layer boundary markers:

{\small
\begin{verbatim}
[PACKET_START]
  [FRAME_START] frame.time_relative 1.834
    frame.time_delta 0.002 [FRAME_END]
  [WLAN_START] wlan.fc.type Data
    wlan.fc.subtype QoS Data wlan.seq 16
    wlan.sa 34:f8:e7:0e:68:d9
    wlan.da 6c:6a:77:45:70:6d [WLAN_END]
  [IP_START] ip.src 10.7.40.10
    ip.dst 10.3.152.95 [IP_END]
  [DNS_START] dns.flags.response 1
    dns.qry.name ws-goguardian.pusher.com
    dns.flags.rcode NoError [DNS_END]
[PACKET_END]
\end{verbatim}
}

\subsection{Dataset and Training}
\label{sec:tokenization:dataset}

A network foundation model must support several modes of reasoning.
Auto-regressive queries (``what should the network answer to this client request?'') demand a generative model; encoder-style queries (``which field value is anomalous?'') demand bidirectional~context.
Although specialized training is always preferable, an auto-regressive model can emulate encoder-style responses via fine-tuning, whereas the reverse is far~harder.
We therefore train \ourmethod{} for Causal Language Modeling~(CLM).

\textbf{Addressing dataset bias.}
In 802.11 networks, the AP sends beacons every $\sim$102\,ms across multiple SSIDs and bands.
A single AP supporting 6~SSIDs in 3~bands produces beacons that, over a 10-minute capture, can number $>$100K quasi-identical frames dominating any na\"ive training~set.
Similarly, clients of similar brands may emit the same keepalive messages, and clients in specific failure conditions may repeat the same request indefinitely, skewing the~corpus.

\begin{figure}[htbp]
  \centering
  \begin{subfigure}[t]{0.22\textwidth}
    \centering
    \includegraphics[width=\textwidth]{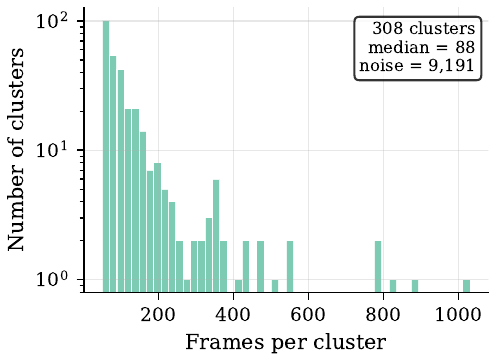}
    \caption{Frame counts per cluster.}
    \label{fig:cluster_dist}
  \end{subfigure}
  \hfill
  \begin{subfigure}[t]{0.22\textwidth}
    \centering
    \includegraphics[width=\textwidth]{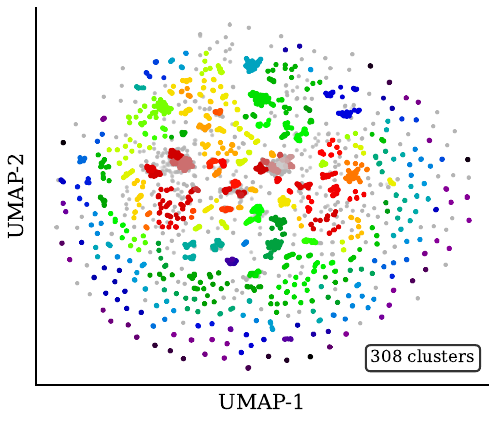}
    \caption{UMAP of embeddings.}
    \label{fig:umap_clusters}
  \end{subfigure}
  \caption{HDBSCAN-based curation. (a)~Long-tail cluster sizes; beacon-dominated clusters contain thousands of near-identical frames. (b)~Frames group by protocol function, validating clustering-based deduplication.}
  \label{fig:clustering}
\end{figure}

We address this through a three-stage curation pipeline.
First, we tokenize each frame and embed it via a generalist embedding model (mxbai-embed-large~\cite{li2023angle}), producing a 1024-dimensional vector capturing both the general intent and internal structure of each~frame.
Second, we project these vectors and apply HDBSCAN~\cite{mcinnes2017hdbscan} clustering, which surfaces $\sim$25K clusters, each made of typical representatives and variants.
As expected, many frames are near-duplicates while others are rare; the Uniform Manifold Approximation and Projection (UMAP)~\cite{mcinnes2018umap} projection in Figure~\ref{fig:umap_clusters} illustrates this distribution.

Third, we apply cosine similarity with MMR~\cite{carbonell1998mmr} to select up to 100 representative samples from each cluster.
For small clusters with fewer than 100~members, we identify varying fields via cosine similarity and generate synthetic members until 100 are~collected.
This reduces beacon dominance from $>$50\% to 4.7\% with 7.6~bits of token entropy (Table~\ref{tab:dataset_stats}).


\section{Evaluation}
\label{sec:eval}

We organize the evaluation as a progressive argument.
We first establish \emph{why} \ourmethod{} works by validating each design lever: architecture choice (\S\ref{sec:eval:architecture}), tokenization (\S\ref{sec:eval:tokenization}), dataset curation (\S\ref{sec:eval:dataset}), and model scaling (\S\ref{sec:eval:scaling}).
We then probe \emph{how deep} the learned representations go via per-field accuracy (\S\ref{sec:eval:perfield}), context-window sensitivity (\S\ref{sec:eval:context}), and cross-category generalization (\S\ref{sec:eval:generalization}).
Next, we ask \emph{how \ourmethod{} compares} to frontier LLMs on the same prediction task (\S\ref{sec:eval:llm}).
Finally, we present \emph{what \ourmethod{} achieves} on three downstream tasks: next-packet prediction (\S\ref{sec:eval:prediction}), zero-shot anomaly detection (\S\ref{sec:eval:anomaly}), and root cause classification (\S\ref{sec:eval:rca}).

\subsection{Experimental Setup}
\label{sec:eval:setup}

\textbf{Model architecture.}
\ourmethod{} uses a GPT-2~\cite{radford2019gpt2} backbone with 12 transformer~\cite{vaswani2017attention} layers, 12 attention heads, and 768-dimensional embeddings, totaling 140M parameters.
The context window is 2,048 tokens.
We train three model sizes sharing the same depth (12 layers) and vocabulary (69K tokens) but differing in width: Small (12H/768D, 140M), Medium (16H/1024D, 225M), and Large (24H/1536D, 450M).
All models are trained from scratch with causal language modeling using AdamW~\cite{loshchilov2019adamw} ($\beta_1{=}0.9$, $\beta_2{=}0.95$), learning rate $7{\times}10^{-4}$ with cosine decay to $7{\times}10^{-5}$, 100 warmup iterations, gradient clipping at 1.0, for 2,000 iterations (20 epochs) with effective batch size 12 ($3 \times 4$ gradient accumulation).

\textbf{Hardware.}
We train and evaluate on an AWS \texttt{g5.\allowbreak{}12xlarge} instance (4$\times$NVIDIA A10G, 48~vCPUs, 192\,GB RAM; \$5.67/hr on-demand), using a single A10G (24\,GB GDDR6, 35~TFLOPS FP32) per run.
Training takes $\sim$6\,h (Small), $\sim$10\,h (Medium), and $\sim$16\,h (Large).
At inference, the Small model occupies 280\,MB in FP16 with 594\,MB peak GPU memory; Medium requires 449\,MB (937\,MB peak) and Large 901\,MB (1,865\,MB peak), all under 8\% of the A10G's 24\,GB VRAM.

\textbf{Training data.}
The training corpus consists of 7,890 PCAP files (149,238 packets, 48.9M tokens) curated via HDBSCAN~\cite{mcinnes2017hdbscan} clustering and MMR~\cite{carbonell1998mmr} sampling from enterprise 802.11 captures.
The validation set contains 2,023 files (38,669 packets, 12.7M tokens).
The vocabulary comprises 69,842 tokens.
All PCAPs are dissected with a single tshark version (4.2.x)~\cite{tshark_manual} to ensure consistent PDML field names; different Wireshark versions may rename or restructure dissector fields, so pinning the version is necessary for reproducibility.

\textbf{Test categories.}
We evaluate on five distinct wireless failure categories from real enterprise deployments:
\begin{sitemize}
\item \textbf{Bad Password} (9,960 files, 116K packets): Authentication failures due to incorrect credentials.
\item \textbf{EAPOL Timeout} (9,992 files, 257K packets): Authentication failures where the exchange times out.
\item \textbf{Invalid PMKID} (2,252 files, 43K packets): Failures from invalid PMKIDs during fast BSS transition.
\item \textbf{Unable to Handle New STA} (9,997 files, 144K packets): AP-side rejections when the station table is full.
\item \textbf{Rejected Temporarily} (9,987 files, 188K packets): Association rejections via transient AP conditions.
\end{sitemize}

For each category, we randomly sample 50 PCAPs and evaluate next-packet prediction: given the first $k$ packets, the model auto-regressively predicts packet $k{+}1$.


\subsection{Architecture and Baselines}
\label{sec:eval:architecture}

Before examining downstream results, we isolate the contribution of the transformer architecture itself.

\textbf{Baselines.}
We compare \ourmethod{} against four baselines:
(1)~\textit{Random}: uniform random token prediction;
(2)~\textit{Most-Frequent}: always predicting the most common token;
(3)~\textit{3-gram}: a trigram language model trained on the same token stream, predicting the most probable next token given the preceding two; and
(4)~\textit{BERT (encoder)}~\cite{devlin2019bert}: a masked language model using the same tokenizer, which can classify but cannot generate or score likelihoods natively.
We do not include netFound~\cite{guthula2023netfound}, Lens~\cite{wang2024lens}, NetGPT~\cite{meng2023netgpt}, or LLMcap~\cite{tulczyjew2024llmcap} as direct baselines because all four target wired or encrypted traffic at the flow or header-byte level and none train on 802.11 management/control frames or use PDML tokenization (Table~\ref{tab:prior_work}).
Retraining them on our corpus would require replacing their tokenizers and input pipelines, reducing the comparison to architecture alone, which is exactly what the GPT vs.\ BERT contrast already~isolates.
Moreover, netFound's 640M-parameter encoder is $4.6\times$ larger than \ourmethod{} yet provides no generation or likelihood-scoring capability; matching its architecture while replacing its tokenizer would test neither netFound's design nor ours, only the shared transformer~backbone.

\begin{table}[t]
\centering
\caption{Design-space comparison of foundation models trained on networking data. MLM~= Masked Language Modeling, CLM~= Causal Language Modeling, Span~= masked span prediction. \ding{51}~= supported, \ding{55}~= not supported, \textbf{--}~= not reported.}
\label{tab:prior_work}
\small
\setlength{\tabcolsep}{3pt}
\begin{tabular}{lccccc}
\toprule
 & \rotatebox{60}{netFound~\cite{guthula2023netfound}} & \rotatebox{60}{Lens~\cite{wang2024lens}} & \rotatebox{60}{NetGPT~\cite{meng2023netgpt}} & \rotatebox{60}{LLMcap~\cite{tulczyjew2024llmcap}} & \rotatebox{60}{\textbf{\ourmethod}} \\
\midrule
Architecture & Enc. & Enc-Dec & Dec. & Enc. & Dec. \\
Parameters & 640M & \textbf{--} & \textbf{--} & ${\sim}$66M & 140M \\
Objective & MLM & Span & CLM & MLM & CLM \\
Tokenization & Field & BPE & BPE & WordPiece & Field \\
Layer hierarchy & Partial & \ding{55} & \ding{55} & \ding{55} & \ding{51} \\
802.11 mgmt/ctrl & \ding{55} & \ding{55} & \ding{55} & \ding{55} & \ding{51} \\
Traffic domain & Wired & Wired & Wired & Telecom & 802.11 \\
Generation & \ding{55} & \ding{51} & \ding{51} & \ding{55} & \ding{51} \\
\bottomrule
\end{tabular}
\end{table}

\begin{table}[t]
\centering
\caption{Architecture and baseline comparison.}
\label{tab:architecture}
\small
\begin{tabular}{lcccc}
\toprule
Model & Token Acc. & Gen? & Anomaly? & Classify? \\
\midrule
\ourmethod (GPT) & 0.831 & \checkmark & \checkmark & \checkmark \\
3-gram & 0.257 & \checkmark & -- & -- \\
Most-Frequent & 0.226 & -- & -- & -- \\
Random & 0.000 & \checkmark & -- & -- \\
BERT (encoder) & -- & -- & -- & \checkmark \\
\bottomrule
\end{tabular}
\end{table}

\ourmethod{} achieves $3.2\times$ the accuracy of the 3-gram baseline and $3.7\times$ that of Most-Frequent (Table~\ref{tab:architecture}), confirming that the transformer captures long-range protocol dependencies that n-gram models cannot.
The 3-gram model's 25.7\% accuracy is only marginally above Most-Frequent (22.6\%), indicating that local bigram context provides little predictive power for protocol field sequences.
\ourmethod{} is the only architecture supporting generation, anomaly detection, and classification from a single next token prediction~objective.
The 83.1\% accuracy here uses a held-out sample ($n{=}405$ predictions) for controlled baseline comparison; per-category evaluation in Table~\ref{tab:prediction_quality} reports category-specific accuracy (74.1--97.3\%) on larger test sets with more~context.

\textbf{Why decoder-only?}
The causal attention mask naturally models the temporal ordering of packet exchanges: request then response, challenge then reply. This provides a better inductive bias for protocol conversations than BERT's~\cite{devlin2019bert} bidirectional attention, which sees future tokens during training.
ELECTRA~\cite{clark2020electra} offers an appealing middle ground (replaced-token detection), but requires a separate generator and cannot natively generate or score arbitrary~sequences.
The auto-regressive factorization provides per-token probabilities for free, enabling zero-shot anomaly detection, generation, and classification from a single objective perspective.

\subsection{Tokenization Ablation}
\label{sec:eval:tokenization}

With the architecture established, we turn to the input representation.
Protocol-aware tokenization yields $6.2\times$ shorter sequences than BPE with higher per-token entropy.
Table~\ref{tab:tokenization_ablation} compares five different tokenization strategies on the same PCAP corpus.

\begin{table}[t]
\centering
\caption{Tokenization ablation: \ourmethod's protocol-aware tokenizer vs.\ alternatives. Compression ratio is relative to byte-level (higher is better).}
\label{tab:tokenization_ablation}
\small
\begin{tabular}{lrrrr}
\toprule
Tokenizer & Vocab & Avg Tok/Pkt & Entropy & Ratio \\
\midrule
\ourmethod (field-value) & 69,842 & 326.8 & 7.61 & 16.2$\times$ \\
BPE on PDML & 100,277 & 2,014.0 & 6.70 & 2.6$\times$ \\
Byte-level & 256 & 5,294.7 & 4.75 & 1.0$\times$ \\
Flat (no layers) & 1,959 & 332.9 & 7.55 & 15.9$\times$ \\
NetGPT-style & 100,277 & 2,013.0 & 6.70 & 2.6$\times$ \\
\bottomrule
\end{tabular}
\end{table}

\begin{figure}[t]
  \centering
  \includegraphics[width=0.90\columnwidth]{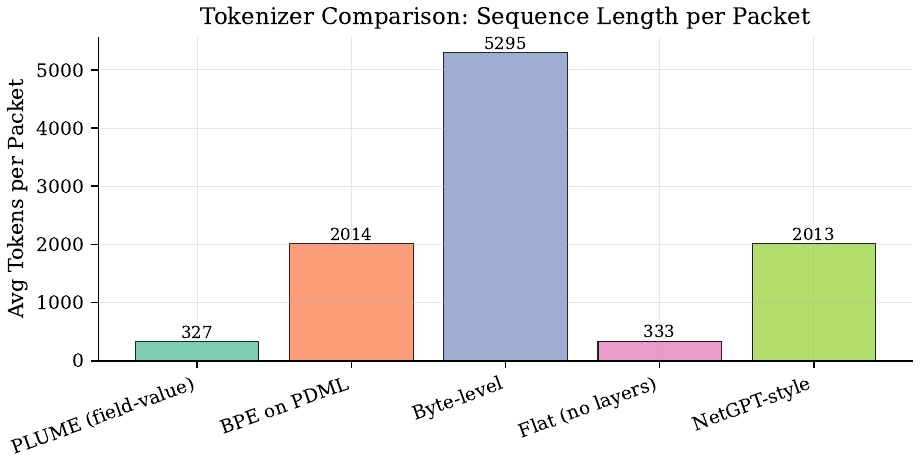}
  \caption{Average tokens per packet. \ourmethod's protocol-aware tokenizer yields $6.2\times$ shorter sequences than BPE and $16.2\times$ shorter than byte-level.}
  \label{fig:tokenizer_comparison}
\end{figure}

\textbf{Sequence length.}
Figure~\ref{fig:tokenizer_comparison} visualizes the difference.
\ourmethod's field-value tokenizer produces $6.2\times$ fewer tokens per packet than BPE and $16.2\times$ fewer than byte-level encoding.
The Flat tokenizer (field-value without layer markers) achieves similar compression but sacrifices the protocol hierarchy that enables the model to distinguish an 802.11 Layer~2 ACK from a TCP Layer~4 ACK.

\textbf{Information density.}
\ourmethod{} achieves the highest per-token entropy (7.61 bits), meaning each token carries maximal information.
BPE and NetGPT-style tokenizers, despite larger vocabularies (100K), achieve only 6.70 bits; byte-level tokenization has the lowest entropy (4.75 bits).

\textbf{Practical implications.}
With a 2,048-token context window, \ourmethod{} can process $\sim$6 complete packets per forward pass, compared to $\sim$1 for BPE and $<$0.4 for byte-level, directly impacting the model's ability to learn cross-packet~patterns.

\subsection{Dataset Quality Ablation}
\label{sec:eval:dataset}

Good tokens require good data.
Table~\ref{tab:dataset_stats} summarizes the curated dataset statistics.
The HDBSCAN~\cite{mcinnes2017hdbscan}+MMR~\cite{carbonell1998mmr} pipeline surfaces $\sim$25K clusters and selects up to 100 representative samples per cluster via cosine similarity with MMR sampling.

\begin{table}[t]
\centering
\caption{Dataset statistics after HDBSCAN+MMR curation. Beacon fraction drops from $>$50\% in raw captures to 4.7\%, while token entropy remains high (7.6 bits).}
\label{tab:dataset_stats}
\small
\begin{tabular}{lrr}
\toprule
Metric & Train & Val \\
\midrule
Files & 7,890 & 2,023 \\
Packets & 149,238 & 38,669 \\
Total tokens & 48.9M & 12.7M \\
Unique tokens & 58,214 & 22,198 \\
Avg packet length (tokens) & 326.8 & 326.8 \\
Token entropy (bits) & 7.601 & 7.610 \\
Beacon fraction & \multicolumn{2}{c}{4.7\% (vs.\ $>$50\% raw)} \\
\bottomrule
\end{tabular}
\end{table}

Table~\ref{tab:test_category_stats} provides per-test-category corpus statistics; packet counts and average lengths vary substantially, reflecting different protocol structures across failure modes.

\begin{table}[t]
\centering
\caption{Per-test-category corpus statistics showing the diversity of the evaluation data.}
\label{tab:test_category_stats}
\small
\begin{tabular}{lrrrr}
\toprule
Category & Files & Packets & Avg Len & Entropy \\
\midrule
Bad Password & 9,960 & 116,469 & 421.8 & 7.52 \\
EAPOL Timeout & 9,992 & 257,057 & 318.9 & 7.37 \\
Invalid PMKID & 2,252 & 43,224 & 387.5 & 7.67 \\
Unable Handle STA & 9,997 & 143,746 & 377.8 & 7.53 \\
Rejected Temp. & 9,987 & 187,871 & 358.3 & 7.58 \\
\bottomrule
\end{tabular}
\end{table}

Near-identical entropy and average packet length across splits (Table~\ref{tab:dataset_stats}) confirm a well-balanced split without information~leakage.
Curated data achieves higher token entropy and lower repetition than raw captures, consistent with compute-optimal scaling findings that data quality matters as much as quantity~\cite{hoffmann2022chinchilla,lee2022dedup}.

\subsection{Multi-Model Scaling and Efficiency}
\label{sec:eval:scaling}

Architecture, tokenizer, and data are now fixed; we vary only model width to find the compute-optimal operating point.
We compare three model widths trained on the same data with the same vocabulary (69K tokens) and depth (12 layers): Small (12H/768D, 140M), Medium (16H/1024D, 225M), and Large (24H/1536D, 450M).
Medium outperforms both alternatives (Table~\ref{tab:scaling}).
At 92.7\% overall token accuracy it edges out Small (90.3\%) and substantially beats Large (80.1\%).
The drop at 450M parameters suggests overfitting given the 48.9M-token corpus, consistent with compute-optimal scaling laws~\cite{hoffmann2022chinchilla}. Prior work \cite{kaplan2020scalinglawsneurallanguage} also shows that performance penalties arise when the ratio between model parameters and dataset size becomes imbalanced, which explains the performance decline for the large model.
All three models share the same tokenizer and data, so these differences isolate the effect of model~width.

\begin{table}[t]
\centering
\caption{Multi-model scaling: same 69K vocabulary and 12-layer depth; differences reflect model width.}
\label{tab:scaling}
\small
\begin{tabular}{lrrrr}
\toprule
Model & Params & Width & Token Acc. & Field Acc. \\
\midrule
Small (12H/768D) & 140M & 768 & 0.903 & 0.884 \\
Medium (16H/1024D) & 225M & 1024 & \textbf{0.927} & \textbf{0.922} \\
Large (24H/1536D) & 450M & 1536 & 0.801 & 0.780 \\
\bottomrule
\end{tabular}
\end{table}

\begin{figure}[t]
  \centering
  \includegraphics[width=\columnwidth]{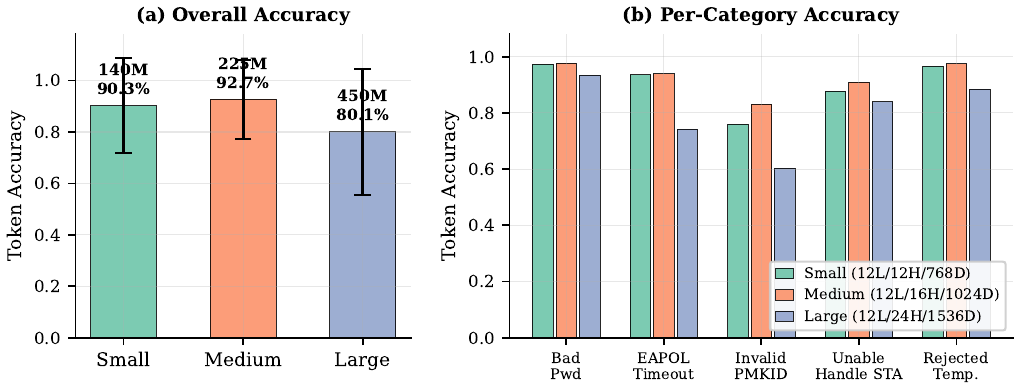}
  \caption{Token accuracy by category for Small (140M), Medium (225M), and Large (450M). Same vocabulary and depth; differences reflect model width.}
  \label{fig:multi_model_scaling}
\end{figure}

Figure~\ref{fig:multi_model_scaling} visualizes the per-category accuracy breakdown.

\textbf{Efficiency.}
Table~\ref{tab:efficiency} summarizes deployment characteristics, benchmarked on a single NVIDIA A10G (24\,GB VRAM).
The Small model achieves the highest throughput ($\sim$200~pkt/s) at under 3\% VRAM; even the largest variant uses under 8\%, leaving ample headroom for larger batches or concurrent~inference.
The marginal cost per packet is effectively zero for all sizes, versus \$4.92 per 1K~packets for GPT-5.2~\cite{openai_gpt52_2025} API calls, representing a $>$500$\times$ cost advantage at~scale.

\begin{table}[t]
\centering
\caption{Efficiency (single A10G, 24\,GB) vs.\ GPT-5.2 API (\$4.92/1K pkt at published pricing~\cite{openai_gpt52_2025}). Peak VRAM includes PyTorch~\cite{paszke2019pytorch} overhead.}
\label{tab:efficiency}
\small
\begin{tabular}{lcccc}
\toprule
 & Small & Medium & Large & GPT-5.2 \\
\midrule
Parameters & 140M & 225M & 450M & N/A \\
Size (FP16) & 280 MB & 449 MB & 901 MB & N/A \\
Peak VRAM & 594 MB & 937 MB & 1,865 MB & N/A \\
Latency (mean) & 280 ms & 384 ms & 666 ms & $\sim$2 s \\
Throughput & 200.2 pkt/s & 150.6 pkt/s & 82.7 pkt/s & N/A \\
Cost/1K pkt & $\sim$\$0 & $\sim$\$0 & $\sim$\$0 & \$4.92 \\
\bottomrule
\end{tabular}
\end{table}


\subsection{Per-Field Micro-Benchmark}
\label{sec:eval:perfield}

The preceding sections show that the right architecture, tokenizer, data, and model width produce strong aggregate accuracy.
We now ask: \emph{which} fields does the model predict well, and where does it struggle?

\begin{figure}[t]
  \centering
  \includegraphics[width=\columnwidth]{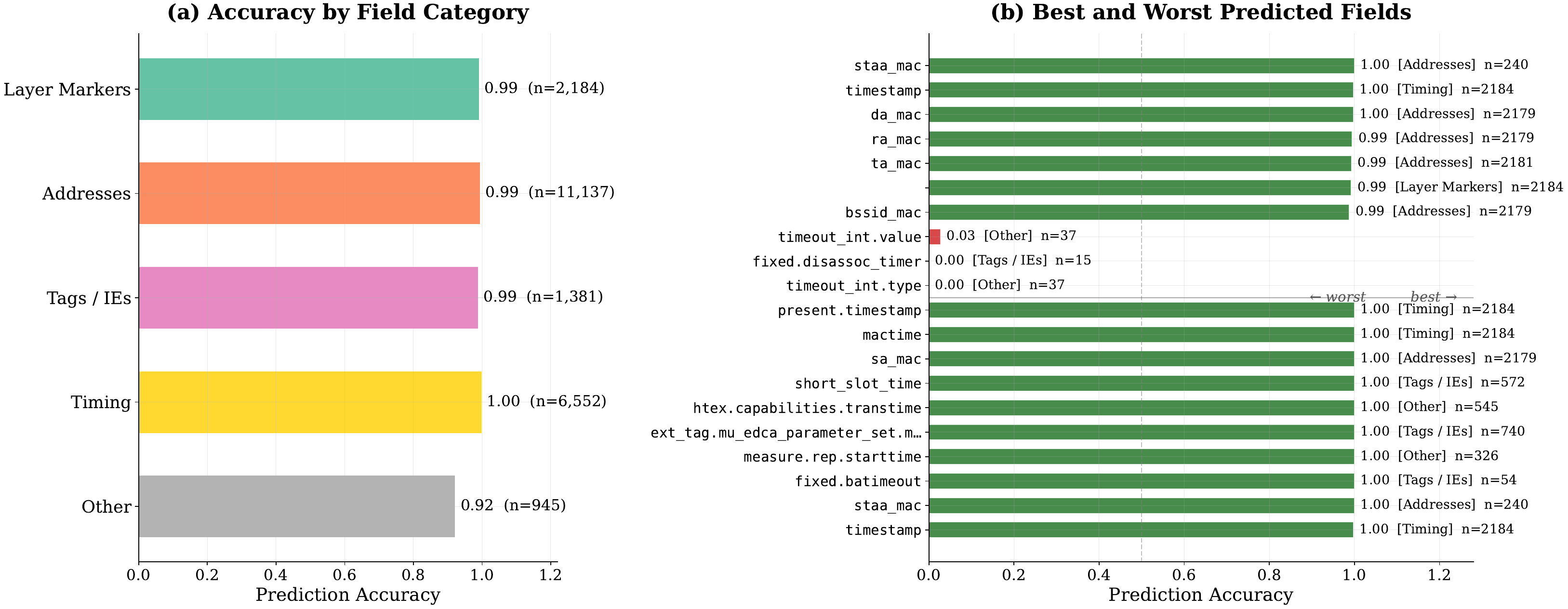}
  \caption{Per-field accuracy by category (left) and 10 best/worst fields (right). Addresses and frame control are near-perfect; timing and rare fields are hardest.}
  \label{fig:per_field}
\end{figure}

Address and frame-control fields achieve near-perfect accuracy; timing and rare fields are hardest to~predict.
We break down prediction accuracy by individual protocol fields across 18 unique fields and 22,199 total predictions (Figure~\ref{fig:per_field}).

Address fields (\texttt{wlan.ra\_mac}, \texttt{wlan.da\_mac}, \texttt{wlan.ta\_mac}, \texttt{wlan.sa\_mac}) achieve perfect or near-perfect accuracy (100\%): in a two-party 802.11 exchange, the responder's addresses are determined by the requester's, so 100\% reflects learned dialogue structure rather than memorization of specific MACs.
Frame control fields and layer markers also exceed 99\%.
Tags and IE fields show moderate accuracy, while numerical values (timing, signal strength) and rare fields are hardest to predict owing to their continuous or low-frequency~nature.

\subsection{Context Window Sensitivity}
\label{sec:eval:context}

Per-field analysis reveals what the model learns; context-window sensitivity reveals how quickly it learns it from preceding packets.

\begin{figure}[t]
  \centering
  \includegraphics[width=0.90\columnwidth]{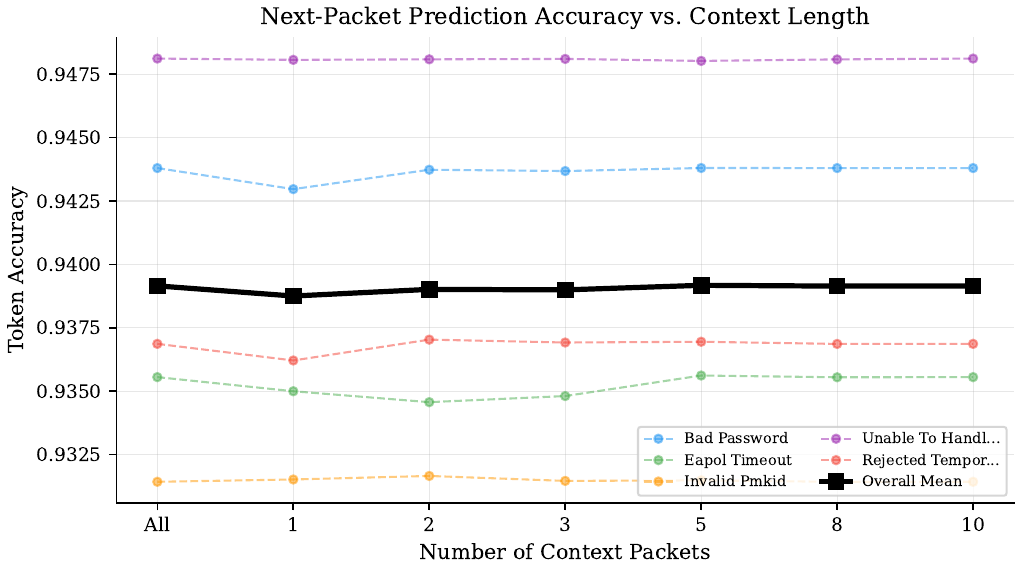}
  \caption{Prediction accuracy vs.\ context length. Accuracy saturates by 2--3 packets, matching typical 802.11 exchange length.}
  \label{fig:context_window}
\end{figure}

Prediction accuracy saturates at 2--3 context packets (Figure~\ref{fig:context_window}).
Token accuracy already exceeds 93\% with just one packet of context and plateaus by two~packets.
This rapid saturation confirms that \ourmethod's protocol-aware tokenization captures sufficient cross-packet state within a small context~window.
The plateau at 3--5 packets aligns with the typical length of 802.11 exchanges.

\subsection{Cross-Category Generalization}
\label{sec:eval:generalization}

The model learns specific fields quickly from minimal context; does this knowledge transfer across failure modes it was never explicitly trained on?

\begin{figure}[t]
  \centering
  \includegraphics[width=\columnwidth]{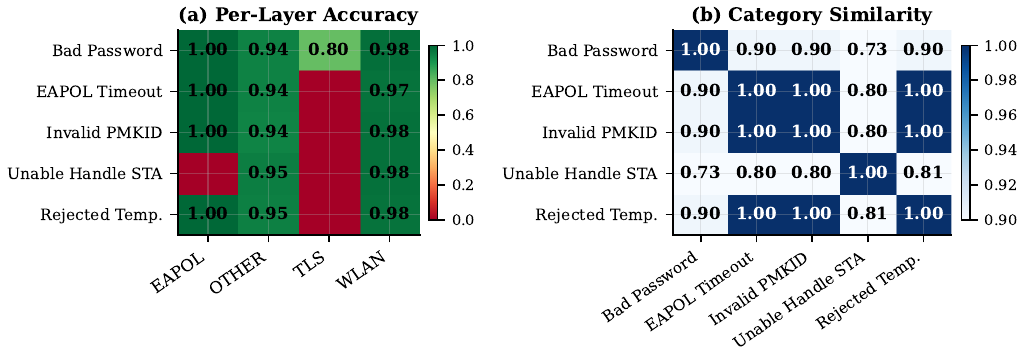}
  \caption{Cross-category generalization. (a)~Per-layer accuracy: WLAN $>$97.5\% across all categories. (b)~Cosine similarity of accuracy profiles: categories with shared protocol flows cluster together.}%
  \label{fig:cross_category}
\end{figure}

WLAN-layer accuracy exceeds 97.5\% across all five failure categories (Figure~\ref{fig:cross_category}), confirming that \ourmethod{} learns general 802.11 structure that transfers across failure modes.
This headline number is dominated by address and frame-control fields that are near-perfectly predictable from conversation context (Figure~\ref{fig:per_field}); timing and rare fields remain the primary source of prediction~error.
We train on a general 802.11 corpus, not on any specific failure category, and evaluate per-layer accuracy.

EAPOL-layer accuracy reaches 100\% in the three categories containing EAPOL frames.
The OTHER layer shows moderate variation (93.3--95.0\% across categories) because it encompasses only the unencrypted Layer~3/4 metadata (IP, ARP, DNS, DHCP headers) present in the PDML dissection; encrypted data-plane payloads are opaque to the dissector and therefore absent from the token~stream.
This is a deliberate scope boundary: \ourmethod{} operates strictly on protocol metadata visible to tshark, so encrypted user traffic is never modeled or leaked, a privacy-by-design property consistent with the on-prem deployment model (\S\ref{sec:usecases}).
The pairwise category similarity matrix reveals two natural clusters: (1)~Bad Password and Unable to Handle New STA (cosine similarity 1.00), sharing similar WLAN-only flows; and (2)~EAPOL Timeout, Invalid PMKID, and Rejected Temporarily (pairwise similarity $>$0.99), all involving EAPOL exchanges.

\subsection{Frontier LLM Comparison}
\label{sec:eval:llm}

\ourmethod{} generalizes across fields, contexts, and categories with a 140M-parameter model.
The natural question is whether frontier LLMs, with orders-of-magnitude more parameters and broader pretraining, can simply be prompted to match this performance.

\begin{figure}[t]
  \centering
  \includegraphics[width=0.90\columnwidth]{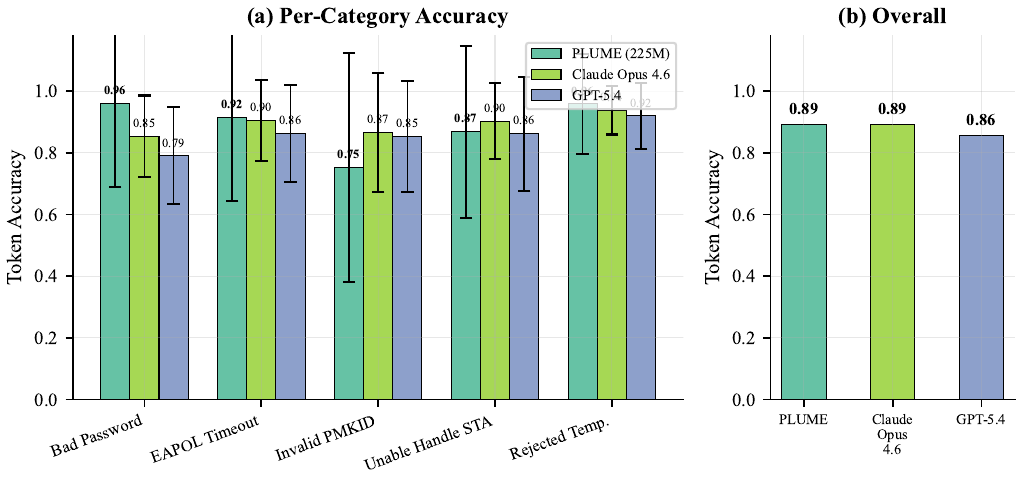}
  \caption{Per-category token accuracy for \ourmethod{} (Small, 140M) vs.\ Claude Opus~4.6 and GPT-5.4. \ourmethod{} matches or exceeds frontier LLMs on stereotyped protocol flows while using $>$600$\times$ fewer parameters.}
  \label{fig:llm_comparison}
\end{figure}

We compare the Small model (140M parameters) against Claude Opus~4.6~\cite{anthropic_claude46_2025} (via AWS Bedrock) and GPT-5.4~\cite{openai_gpt54_2025} (via Azure AI Foundry) on the same next-packet prediction task.
Each LLM receives identical tokenized context and generates a free-form completion aligned to the ground-truth token sequence for scoring. \ourmethod{} leads on three of five categories (Table~\ref{tab:llm_comparison}, Figure~\ref{fig:llm_comparison}), with the largest margin on stereotyped authentication flows such as Bad Password, where protocol-specific inductive bias dominates.
Claude edges ahead on Invalid PMKID and Unable to Handle New STA, both involving diverse key-negotiation or rejection patterns where broader world knowledge helps.
Overall means are comparable (\ourmethod{} 89.1\%, Claude 89.3\%, GPT-5.4 85.7\%), but \ourmethod{} achieves this at $>$600$\times$ fewer parameters and effectively zero marginal cost on a single GPU.

\begin{table}[t]
\centering
\caption{Next-packet token accuracy: \ourmethod{} (Small, 140M) vs.\ frontier LLMs. Total of 682 prediction pairs across five categories. Best per category in \textbf{bold}.}
\label{tab:llm_comparison}
\small
\begin{tabular}{lccc}
\toprule
Failure Category & \ourmethod & Claude 4.6 & GPT-5.4 \\
\midrule
Bad Password & \textbf{0.960} & 0.854 & 0.790 \\
EAPOL Timeout & \textbf{0.915} & 0.904 & 0.863 \\
Invalid PMKID & 0.753 & \textbf{0.865} & 0.852 \\
Unable to Handle New STA & 0.868 & \textbf{0.903} & 0.862 \\
Rejected Temporarily & \textbf{0.958} & 0.938 & 0.920 \\
\midrule
Overall mean & 0.891 & 0.893 & 0.857 \\
\bottomrule
\end{tabular}
\end{table}

\textbf{Why does \ourmethod{} win on stereotyped flows?}
Bad Password and Rejected Temporarily follow narrow, predictable protocol sequences where field-level tokenization and causal pretraining on 802.11 traces provide a strong inductive bias.
Frontier LLMs process the same fields as flat text tokens and lack the protocol-aware vocabulary that lets \ourmethod{} predict entire field-value pairs in a single step.

\textbf{Where do LLMs help?}
Invalid PMKID involves diverse Fast BSS Transition (FT) patterns where valid Robust Security Network Element (RSNE), Mobility Domain IE (MDIE), and Fast Transition IE (FTIE) combinations depend on the AP's Protected Management Frames (PMF) and FT policies.
Frontier LLMs can partially recover these constraints from broad pretraining on 802.11 specification text, whereas \ourmethod{} must infer them solely from observed token co-occurrences.
This suggests that hybrid approaches, using \ourmethod{} for structured protocol fields and an LLM for rare or policy-dependent content, could combine the strengths of both.


\subsection{Next-Packet Prediction Quality}
\label{sec:eval:prediction}

Having established why \ourmethod{} works and how it compares to alternatives, we now present the full downstream results.

\ourmethod{} achieves 74.1--97.3\% token accuracy across five failure categories (Table~\ref{tab:prediction_quality}).
Field accuracy is consistently above 95\% in all categories despite the wide spread in token accuracy.
Bad Password is highest (97.3\%) because authentication-failure sequences follow a narrow, predictable protocol flow.
Invalid PMKID is lowest (74.1\%) because fast-BSS-transition sequences involve diverse key negotiation patterns; the wide confidence interval (0.658--0.822) reflects a bimodal split where PCAPs with standard PMKID renegotiation score above 0.90, while those with non-standard RSNE/MDIE/FTIE combinations in multi-AP roaming scenarios fall below 0.20.
Perplexity remains low (2.1--2.3) across all categories, confirming that the model assigns high probability to correct next~tokens.

\begin{table}[t]
\centering
\caption{Per-category prediction quality (50 PCAPs each; $\pm$ denotes std across PCAPs).}
\label{tab:prediction_quality}
\small
\begin{tabular}{lcccc}
\toprule
Failure Category & Token Acc. & Field Acc. & PPL & $n$ \\
\midrule
Bad Password & 0.973 $\pm$ 0.110 & 0.963 & 2.1 & 50 \\
EAPOL Timeout & 0.937 $\pm$ 0.182 & 0.959 & 2.3 & 50 \\
Invalid PMKID & 0.741 $\pm$ 0.299 & 0.957 & 2.3 & 50 \\
Unable to Handle. STA & 0.880 $\pm$ 0.111 & 0.963 & 2.1 & 50 \\
Rejected Temporarily & 0.969 $\pm$ 0.063 & 0.962 & 2.1 & 50 \\
\bottomrule
\end{tabular}
\end{table}

\begin{figure}[t]
  \centering
  \includegraphics[width=0.85\columnwidth]{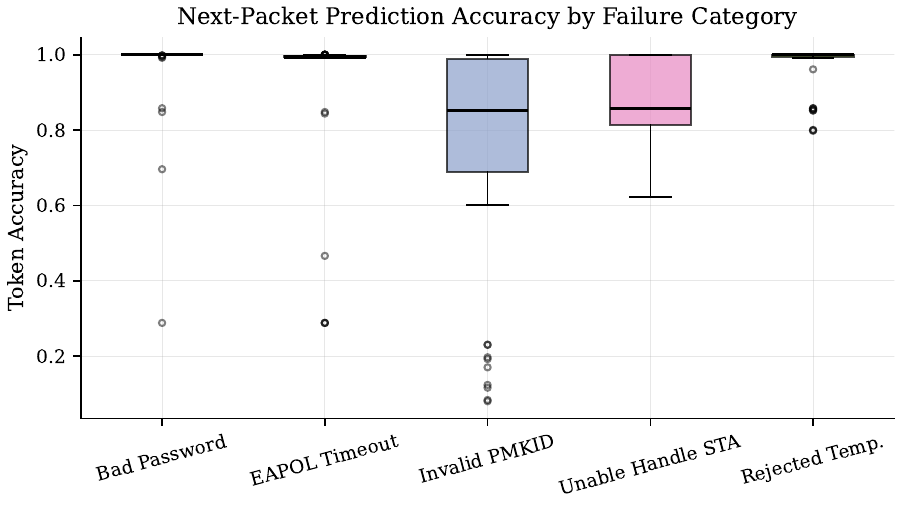}
  \caption{Per-PCAP token accuracy distribution ($n{=}50$). Boxes: IQR; whiskers: $1.5\times$IQR; circles: outliers.}
  \label{fig:accuracy_boxplots}
\end{figure}

\textbf{Variance analysis.}
Figure~\ref{fig:accuracy_boxplots} shows the distribution of per-PCAP token accuracy.
Rejected Temporarily exhibits the tightest spread (std$=$0.063), reflecting stereotyped rejection sequences.
Invalid PMKID has the widest (std$=$0.299), consistent with the bimodal split described~above.

\begin{figure}[t]
  \centering
  \includegraphics[width=0.75\columnwidth]{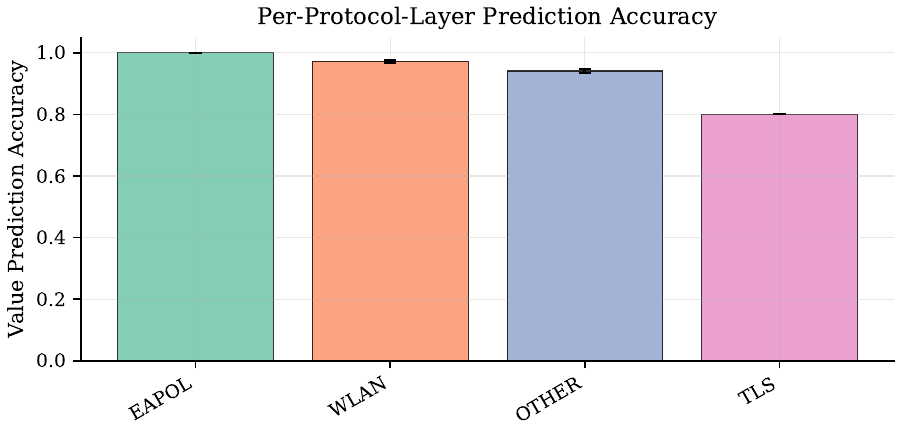}
  \caption{Per-protocol-layer prediction accuracy (mean $\pm$ std across categories). EAPOL fields are predicted perfectly; WLAN exceeds 96\%; OTHER reaches 93--95\%.}
  \label{fig:layer_accuracy}
\end{figure}

\textbf{Per-protocol-layer accuracy.}
Figure~\ref{fig:layer_accuracy} breaks down prediction accuracy by protocol layer.
EAPOL fields are predicted perfectly in all three categories where they appear.
WLAN fields reach 96.5--97.8\% and OTHER fields 93.3--95.0\%, reflecting the model's strong 802.11 inductive bias from training data dominated by wireless management~frames.

\textbf{Bootstrap confidence intervals.}
Table~\ref{tab:bootstrap_ci} reports 95\% bootstrap confidence intervals for token accuracy.
All categories show non-overlapping CIs with the random baseline (0.0\%), confirming robust performance across failure~modes.

\begin{table}[t]
\centering
\caption{95\% bootstrap confidence intervals for per-category token accuracy ($n{=}50$ PCAPs each).}
\label{tab:bootstrap_ci}
\small
\begin{tabular}{lccc}
\toprule
Failure Category & Mean & CI Lower & CI Upper \\
\midrule
Bad Password & 0.973 & 0.942 & 0.996 \\
EAPOL Timeout & 0.937 & 0.886 & 0.977 \\
Invalid PMKID & 0.741 & 0.658 & 0.822 \\
Unable to Handle New STA & 0.880 & 0.849 & 0.912 \\
Rejected Temporarily & 0.969 & 0.952 & 0.986 \\
\bottomrule
\end{tabular}
\end{table}

\subsection{Zero-Shot Anomaly Detection}
\label{sec:eval:anomaly}

The per-token probabilities that drive prediction accuracy also provide a zero-shot anomaly detector, requiring no labeled failure data.

\ourmethod{} achieves AUROC $\geq$0.99 across all five failure categories without any labeled anomaly data.
For each PCAP, we compute the mean per-token probability under the~model.
Healthy captures yield high mean probabilities ($>$0.99), while failure-category captures show systematically lower probabilities; we use this gap to discriminate healthy from failure captures via a simple threshold.
We set a single global threshold at the 5th percentile of mean per-token probability over the healthy validation split (not tuned per category); captures scoring below this threshold are flagged as anomalous.
No failure-category labels or failure-category statistics inform the threshold, preserving the zero-shot~property.

Table~\ref{tab:anomaly_detection} reports AUROC and Area Under the Precision-Recall Curve (AUPRC) for each failure category.
The mean per-token probability gap between healthy and failure captures is sufficient for zero-shot~detection.

\textbf{Statistical baseline.}
A simple packet-length baseline (mean tokens per packet as anomaly score) achieves AUROC 0.95 for Bad Password, where failure PCAPs are substantially longer.
For categories with packet lengths closer to healthy traffic, such as Invalid PMKID (0.68) and EAPOL Timeout (0.78), the statistical baseline degrades sharply, while \ourmethod{} maintains AUROC $\geq$0.99 across all five categories (Table~\ref{tab:anomaly_detection}).

\begin{table}[t]
\centering
\caption{Zero-shot anomaly detection (AUROC/AUPRC, $n{=}50$ per category vs. healthy baselines).}
\label{tab:anomaly_detection}
\small
\begin{tabular}{lcccc}
\toprule
Failure Category & \multicolumn{2}{c}{\ourmethod} & \multicolumn{2}{c}{Pkt-Length} \\
\cmidrule(lr){2-3} \cmidrule(lr){4-5}
 & AUROC & AUPRC & AUROC & AUPRC \\
\midrule
Bad Password & 0.99 & 0.99 & 0.95 & 0.95 \\
EAPOL Timeout & 1.00 & 1.00 & 0.78 & 0.76 \\
Invalid PMKID & 1.00 & 1.00 & 0.68 & 0.75 \\
Unable to Handle. STA & 1.00 & 1.00 & 0.79 & 0.80 \\
Rejected Temporarily & 1.00 & 1.00 & 0.79 & 0.73 \\
\bottomrule
\end{tabular}
\end{table}

\textbf{Why AUROC$\geq$0.99 is not data leakage.}
The model is trained exclusively on healthy 802.11 captures and never sees failure-category labels.
The near-perfect separation reflects the rigidity of the 802.11 state machine: protocol failures (e.g., a Deauthentication where EAPOL Message~3 was expected, or a missing ACK after Association Response) are \emph{syntactically} anomalous relative to the well-defined handshake grammar the model learns during pretraining.
Deviations from rigid protocol syntax produce low per-token probabilities by construction, analogous to how language models trained on valid source code trivially flag syntax~errors.

\begin{figure}[t]
  \centering
  \includegraphics[width=0.90\columnwidth]{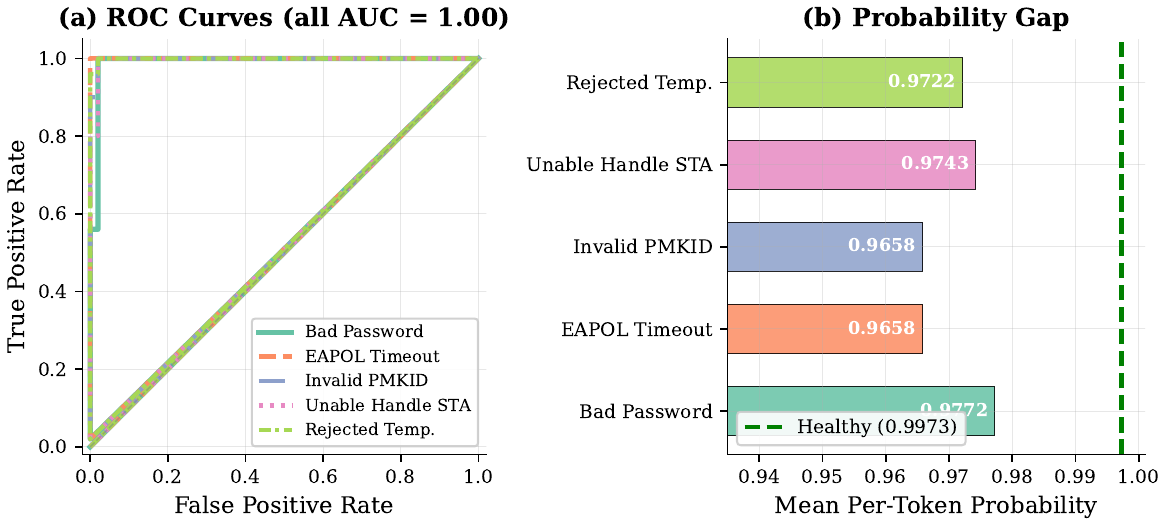}
  \caption{Zero-shot anomaly detection. (a)~ROC curves: AUROC$\geq$0.99 for all categories. (b)~Mean per-token probability vs.\ healthy baseline (dashed).}
  \label{fig:roc_auroc}
\end{figure}

\begin{figure}[t]
  \centering
  \includegraphics[width=0.75\columnwidth]{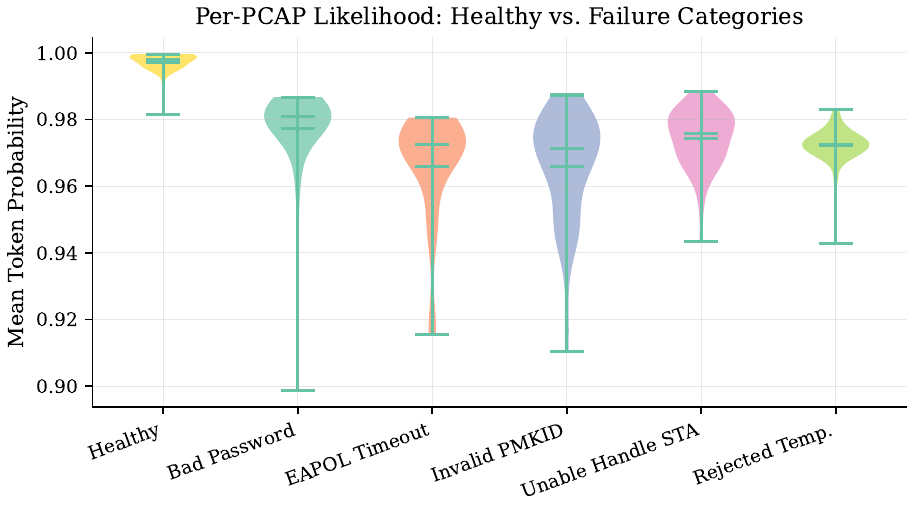}
  \caption{Per-token probability: healthy (green) vs.\ failure (red). Failure captures consistently score lower.}
  \label{fig:kl_violin}
\end{figure}

Figure~\ref{fig:roc_auroc} shows the per-category ROC curves and probability gap; Figure~\ref{fig:kl_violin} plots the full per-token probability distribution confirming the clear separation.

\subsection{Root Cause Classification}
\label{sec:eval:rca}

Anomaly detection flags \emph{that} something is wrong; root cause classification identifies \emph{what}. \ourmethod's unsupervised features carry discriminative signal for root cause classification.
We extract a 19-dimensional feature vector per PCAP (mean probability, standard deviation, median, log-probability, and rank statistics) and train lightweight classifiers to discriminate the five failure categories.
A Random Forest achieves 73.2\% five-class accuracy, $3.7\times$ the 20\% random baseline, without any task-specific training (Table~\ref{tab:rca_classification}).

\begin{table}[t]
\centering
\caption{Root cause classification: three classifiers on 19-D per-PCAP features ($n{=}250$, 5 categories $\times$ 50).}
\label{tab:rca_classification}
\small
\begin{tabular}{lccc}
\toprule
Classifier & Accuracy & Macro F1 & Best Category \\
\midrule
Logistic Regression & 0.68 & 0.68 & Rejected Temp. (0.83) \\
Random Forest & 0.73 & 0.73 & Rejected Temp. (0.85) \\
kNN & 0.62 & 0.61 & Rejected Temp. (0.78) \\
\bottomrule
\end{tabular}
\end{table}

\begin{figure}[t]
  \centering
  \includegraphics[width=0.95\columnwidth]{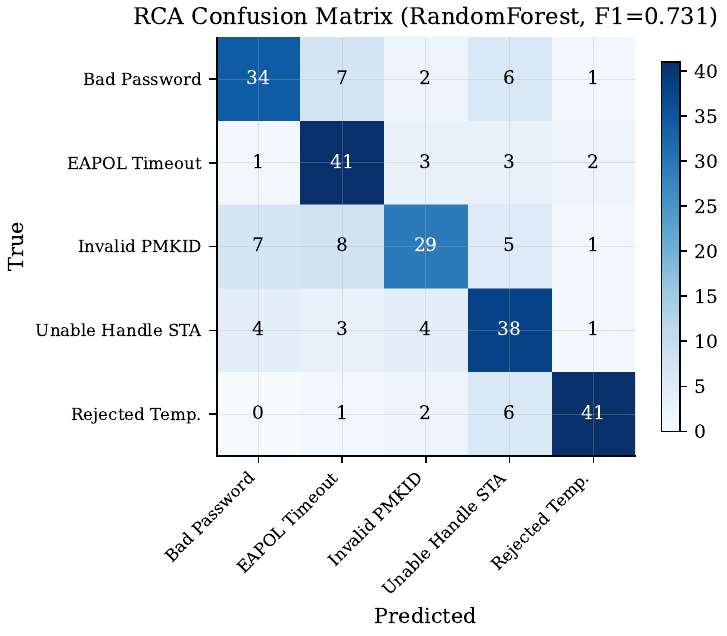}
  \caption{Logistic Regression confusion matrix ($n{=}250$). Most reliable: Rejected Temporarily, Unable to Handle New STA. Most confused: Invalid PMKID.}
  \label{fig:confusion_matrix}
\end{figure}

Figure~\ref{fig:confusion_matrix} shows the confusion matrix for Logistic Regression.
Rejected Temporarily and Unable to Handle New STA are classified most reliably; Invalid PMKID is most often confused with Bad Password and EAPOL Timeout, consistent with overlapping EAPOL-layer protocol flows. The confusion patterns align with protocol structure: categories sharing EAPOL-layer flows (EAPOL Timeout, Invalid PMKID) are confused with each other, while Rejected Temporarily and Unable to Handle New STA occupy more distinct features.

\textbf{Practical utility.}
At 73.2\% five-class accuracy, the classifier serves as a triage tool rather than a definitive diagnosis.
Rejected Temporarily (F1$=$0.85) and Unable to Handle New STA are reliably separated and can trigger targeted remediation (Table~\ref{tab:rca_classification}); Invalid PMKID and EAPOL Timeout, which share EAPOL-layer flows, are frequently confused and require human follow-up.
Even for confused categories, narrowing to two or three reduces the time to investigate.

\begin{figure}[t]
  \centering
  \includegraphics[width=0.80\columnwidth]{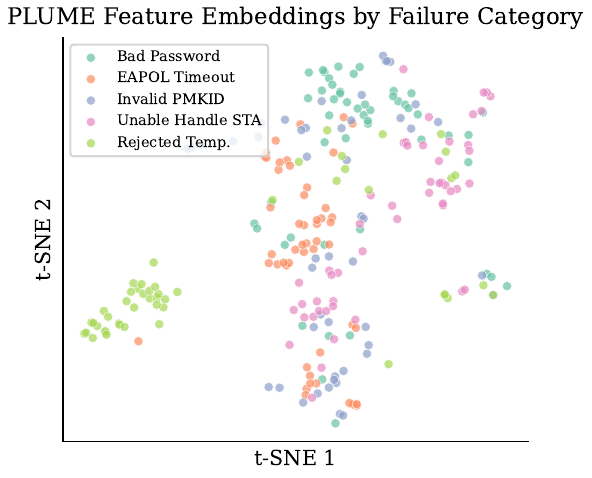}
  \caption{t-SNE of 19-D per-PCAP features colored by failure category. EAPOL-related categories partially overlap, reflecting shared protocol flows.}
  \label{fig:tsne_clusters}
\end{figure}

Figure~\ref{fig:tsne_clusters} provides a complementary view via t-SNE projection of the 19-dimensional feature vectors.
The five categories form visually separable clusters, with Bad Password and Unable to Handle New STA occupying distinct regions.
EAPOL Timeout, Invalid PMKID, and Rejected Temporarily show partial overlap, consistent with shared EAPOL-layer flows and the confusion patterns in Figure~\ref{fig:confusion_matrix}.

\section{Use Cases and System Integration}
\label{sec:usecases}

\ourmethod{} is designed not as a standalone classifier but as a \emph{callable tool} within larger diagnostic~workflows.
Two concrete use cases leverage its combination of generation, likelihood scoring, and protocol-native representations.

\subsection{Multi-Agent RCA}
\label{sec:usecases:rca}

In a multi-agent RCA architecture, a planner agent (e.g., an LLM orchestrator) receives a user complaint and coordinates specialized~tools.
\ourmethod{} serves as the packet analysis tool:
(1)~the planner retrieves the relevant PCAP slice and passes it to \ourmethod{} in tokenized PDML form;
(2)~\ourmethod{} processes the sequence auto-regressively, flagging tokens with anomalously low likelihood;
(3)~it generates the \emph{expected} next packet, and divergence between expected and actual localizes the failure~point;
(4)~the planner receives a structured summary, e.g., ``EAPOL Message~3 expected after Message~2 (seq=4), but AP sent Deauthentication (reason=2). Likely cause: PMKID mismatch or PMF policy conflict.''

\begin{enumerate}
\item The planner retrieves the relevant PCAP slice (e.g., the 30-second window around the failure event) and passes it to \ourmethod{} in tokenized PDML form.
\item \ourmethod{} processes the sequence auto-regressively, producing per-token log-likelihoods. Tokens with anomalously low likelihood flag unexpected field values or missing protocol~steps.
\item \ourmethod{} generates the \emph{expected} next packet given the conversation so far. Divergence between expected and actual packets localizes the failure~point.
\item The planner receives a structured summary: ``EAPOL Message~3 expected after Message~2 (seq=4), but AP sent Deauthentication (reason=2, `previously authenticated STA leaving'). Likely cause: PMKID mismatch or PMF policy conflict.''
\end{enumerate}

Because \ourmethod{} exchanges \emph{explanations} rather than raw packets, the planner never sees sensitive payload data, only protocol-level summaries. Our evaluation validates this pipeline's foundation.
A Random Forest on 19-dimensional likelihood features achieves 73.2\% five-class root cause accuracy, $3.7\times$ the random baseline, with per-category F1 reaching 0.85 for Rejected Temporarily (\S\ref{sec:eval:rca}).
The confusion matrix (\S\ref{sec:eval:rca}) and t-SNE projection (Figure~\ref{fig:tsne_clusters}) confirm that the five failure categories are distinguishable from representations alone.

\subsection{Proactive Anomaly Detection}
\label{sec:usecases:anomaly}

\ourmethod's auto-regressive likelihood provides a zero-shot anomaly detector without requiring labeled anomaly~data.
For each packet, we compute the average per-token log-likelihood; packets deviating significantly from the model's expectations are flagged as anomalous. 
Our evaluation (\S\ref{sec:eval:anomaly}) confirms this capability.
\ourmethod{} achieves AUROC$\geq$0.99 for all five failure categories without any labeled anomaly training data (Table~\ref{tab:anomaly_detection}).
The mean per-token probability gap between healthy and failure captures is sufficient for reliable zero-shot detection across all failure~modes.
The per-protocol-layer accuracy (Figure~\ref{fig:layer_accuracy}) confirms that anomalies in WLAN-layer are detected.





\section{Related Work}
\label{sec:related}

We organize related work into seven categories and position \ourmethod{} with respect to~each.

\textbf{Foundation models for network traffic.}
Lens~\cite{wang2024lens} pre-trains a transformer via knowledge-guided masked span prediction~\cite{raffel2020t5} for traffic classification and~generation.
netFound~\cite{guthula2023netfound} pre-trains on unlabeled packet traces with self-supervised multi-modal embeddings.
DBF-PSR~\cite{Ding2025DBF-PSR} employs dual-branch fusion with protocol semantic representations for traffic~classification.
These works operate on flow- or header-level features with standard tokenization (BPE~\cite{sennrich2016bpe} or fixed byte chunks) and do not address 802.11-specific data quality challenges, namely beacon dominance, reactive capture bias, and missing pre-failure~context.
\ourmethod{} complements them by centering on PDML dissections, field-boundary tokenization, and curated training~data.

\textbf{Tokenization for structured networking data.}
netFound~\cite{guthula2023netfound} uses field-level tokens without preserving hierarchy or timing; NetGPT~\cite{meng2023netgpt} applies BPE~\cite{sennrich2016bpe} to raw packet text, fragmenting field~boundaries.
The Byte Latent Transformer~\cite{pagnoni2024blt} shows that dynamic, content-aware patching can rival fixed vocabularies at scale; \ourmethod{} aligns patches to protocol field boundaries, yielding 7.61 bits of per-token entropy vs.\ 4.75 for byte-level based system (Table~\ref{tab:tokenization_ablation}).

\textbf{Adapting LLMs for networking tasks.}
NetLLM~\cite{wu2024netllm} adapts pretrained LLMs to networking tasks (viewport prediction, adaptive bit-rate, cluster scheduling) by converting multi-modal data into token sequences with task-specific~heads.
TrafficLLM~\cite{cui2025trafficllm} proposes dual-stage fine-tuning with traffic-domain tokenization, reporting high F1 scores across 229 traffic~types.
Both adapt text-native LLMs; \ourmethod{} instead builds a \emph{network-native} model with a protocol-structure tokenizer, yielding $6.2\times$ shorter sequences at 140M parameters.

\textbf{Generative pretrained models for network traffic.}
NetGPT~\cite{meng2023netgpt} tokenizes multi-pattern traffic into unified text via header-field shuffling, packet segmentation, and prompt~labels.
TrafficGPT~\cite{qu2024trafficgpt} extends this with a linear-attention transformer.
NetDiffusion~\cite{jiang2024netdiffusion} takes a diffusion-based approach, generating protocol-constrained synthetic traces for data~augmentation.
All rely on BPE~\cite{sennrich2016bpe} or byte-level tokenization, inheriting the mismatch we quantify in \S\ref{sec:eval:tokenization}: sequences $6.2\times$ longer than \ourmethod's protocol-aware tokenizer (Table~\ref{tab:tokenization_ablation}).
\ourmethod's field-value tokenization yields shorter, higher-entropy sequences that let the model see more packets per context window and learn denser cross-packet~patterns.

\textbf{LLMs applied to PCAP analysis.}
LLMcap~\cite{tulczyjew2024llmcap} applies masked language modeling to PCAP data, learning the \textit{grammar, context, and structure} of successful captures for unsupervised failure detection, and is the closest work to \ourmethod{} in~spirit.
However, its BERT-style~\cite{devlin2019bert} objective cannot generate next packets or produce per-token likelihoods natively. 
However, LLMcap operates without protocol layer hierarchy and does not address data curation challenges of 802.11~captures.
\ourmethod's auto-regressive objective enables generation, anomaly scoring, and classification from a single model while preserving layer structure (e.g., 802.11 L2 ACK from a TCP L4~ACK).

\textbf{Traffic classification with pretrained transformers.}
ET-BERT~\cite{lin2022etbert} pre-trains deep contextualized datagram-level representations for encrypted traffic~classification.
YaTC~\cite{zhao2023yatc} uses a masked auto-encoder with multi-level flow representation for few-shot~classification.
NetMamba~\cite{wang2024netmamba} replaces the transformer with a unidirectional Mamba~\cite{gu2023mamba} state-space model for linear-time classification. All target encrypted traffic at the flow/datagram level; \ourmethod{} targets 802.11 management and control frames using protocol structure.

\textbf{Data quality for pretraining.}
Compute-optimal scaling~\cite{hoffmann2022chinchilla} shows that data quantity and model size should be~balanced.
Deduplication~\cite{lee2022dedup} curbs memorization and improves~generalization.
RefinedWeb~\cite{penedo2023refinedweb} shows that rigorous web-data filtering and deduplication alone can outperform curated corpora. \ourmethod{} applies these principles to packet traces: HDBSCAN~\cite{mcinnes2017hdbscan} clustering identifies redundant frames, and MMR~\cite{carbonell1998mmr} preserves diversity while eliminating repetition, reducing beacon fraction to 4.7\% (Table~\ref{tab:dataset_stats}).
\section{Conclusion and Future Work}
\label{sec:conclusion}

We presented \ourmethod, a compact 140M-parameter foundation model for 802.11 wireless traces built on three interlocking ideas: a protocol-aware tokenizer that yields $6.2\times$ shorter sequences than BPE, curated training data that suppresses beacon dominance while preserving rare events, and a decoder-only auto-regressive objective that unifies generation, anomaly detection, and classification in a single~model.

Across five real-world failure categories, \ourmethod{} achieves 74--97\% next-packet token accuracy and AUROC$\geq$0.99 for zero-shot anomaly detection.
Head-to-head, it matches or exceeds Claude Opus~4.6~\cite{anthropic_claude46_2025} and GPT-5.4~\cite{openai_gpt54_2025} on the same prediction task (Table~\ref{tab:llm_comparison}) with $>$600$\times$ fewer parameters and effectively zero marginal cost on a single~GPU.

\textbf{Limitations and future work.}
\ourmethod{} currently targets 802.11 management and control frames; extending to data-plane protocols is a natural next~step.
Root cause classification from unsupervised features would benefit from larger evaluation sets and task-specific feature~engineering.
The anomaly detector uses raw per-token probabilities without calibration~\cite{guo2017calibration}, and we do not yet characterize \emph{confidently wrong} predictions, both important for safe automated~RCA.
All data come from a single enterprise deployment; multi-site evaluation and cross-domain benchmarking remain~open.

\ourmethod{} demonstrates that \emph{representation matters}: a small model with the right tokenizer and training data can match or exceed much larger general-purpose models.
We expect this principle to hold broadly as foundation models expand into new structured-data~domains.

\bibliographystyle{ACM-Reference-Format}
\bibliography{lpm}

\end{document}